\documentclass[letterpaper, 10 pt, conference]{ieeeconf}  % Comment this line out if you need a4paper

\IEEEoverridecommandlockouts                              % This command is only needed if 
\usepackage[utf8]{inputenc}
\usepackage{textgreek}
\usepackage{graphicx}
\usepackage{amsmath}
\usepackage{amssymb}
\usepackage{graphicx}
\usepackage{bm}
\usepackage{booktabs} % \toprule, \midrule, \bottomrule
\usepackage{multirow} % \multirow
\usepackage{xcolor}
\usepackage{hyperref}

\usepackage{booktabs,multirow,tabularx,xcolor}

\newcolumntype{Y}{%
    >{\hsize=0.9\hsize\linewidth=\hsize
      \centering\arraybackslash}X%
}
\newcolumntype{Z}{%
    >{\hsize=1.2\hsize\linewidth=\hsize
      \centering\arraybackslash}X%
}

\title{\LARGE \bf
Breaking the Vision–Action Shortcut: Latent Interface Training for Generalizable Robotics Foundation Models
}

\author{Jianman Lin$^{1,*}$, Shailesh Shailesh$^{2,*}$, Zhongyi Luo$^{3}$, Jiafei Duan$^{2}$%
\thanks{$^{*}$Equal co-first contribution.}%
\thanks{$^{1}$Jianman Lin is with the School of Future Technology,
        South China University of Technology, Guangzhou, China
        {\tt\small linjianmancjx@gmail.com}}%
\thanks{$^{2}$Shailesh Shailesh is with the National University of Singapore,
        Singapore
        {\tt\small shailesh.xml@nus.edu.sg}}%
\thanks{$^{3}$Zhongyi Luo is with Nanyang Technological University,
        Singapore
        {\tt\small LUOZ0031@e.ntu.edu.sg}}%
\thanks{$^{2}$Jiafei Duan is with the School of Computing,
        National University of Singapore, Singapore
        {\tt\small duanj1@nus.edu.sg}}%
\thanks{Jiafei Duan is the corresponding author.}%
}

\begin{document}

\maketitle
\thispagestyle{empty}
\pagestyle{empty}

%%%%%%%%%%%%%%%%%%%%%%%%%%%%%%%%%%%%%%%%%%%%%%%%%%%%%%%%%%%%%%%%%%%%%%%%%%%%%%%%
\begin{abstract}
Robot foundation models achieve strong in-distribution performance
but often degrade under visual distribution shifts.
When learning to generate actions from pretrained visual
representations, models may exploit task-irrelevant visual
cues that correlate with demonstrated actions within the
training distribution.
Such \emph{vision--action shortcuts} can undermine generalization
when these correlations change under distribution shifts.
Mitigating these shortcuts requires constraining how visual
information is used for action generation while preserving
task-relevant spatial information.
We propose \textbf{Latent Interface Training (LIT)}, a
framework-agnostic two-stage strategy that first establishes
a spatial-goal-conditioned action prior without images, then
constrains visual conditioning through a pose-supervised
latent interface.
Stage~1 trains the action expert to generate action chunks
conditioned on language, robot state, and each demonstrated
chunk's terminal SE(3) end-effector pose, learning goal-directed
action generation independently of visual cues.
Stage~2 introduces a latent interface that aggregates visual
and semantic representations and serves as the pretrained
action expert's only visual conditioning pathway.
The interface is supervised to reconstruct the terminal pose
previously used to condition Stage~1, encouraging it to retain
the goal-relevant spatial information needed for action generation.
Across four vision--language--action and world--action
architectures---$\pi_{0.5}$, MolmoAct2, FAST-WAM, and ImageWAM---%
LIT improves overall LIBERO-Plus success by 3.87--10.70
percentage points while preserving or improving average
LIBERO success.
Real-world evaluations show 13.30--16.70 percentage-point gains
in success aggregated across three tasks under unseen camera
configurations, lighting variations, and distractors. Project page: \href{https://magiclab-nus.github.io/LIT/?v=37955c5}{\textcolor{blue}{magiclab-nus.github.io/LIT}}
\end{abstract}

\section{INTRODUCTION}
\begin{figure*}[t]
    \centering
    \includegraphics[width=\textwidth]{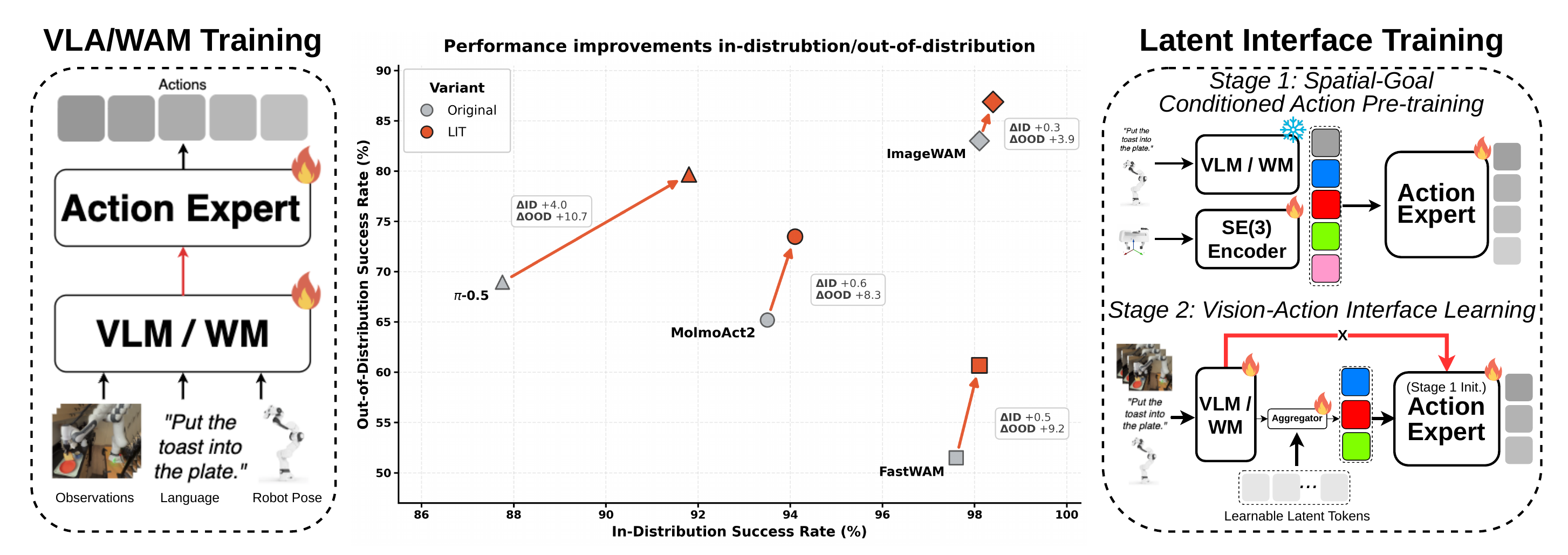}
    \vspace{-20pt}
    \caption{\textbf{Training paradigms and generalization performance.}
    Standard training conditions the action expert on visual representations
    (left). LIT (right) first learns a spatial-goal-conditioned action prior
    without images to avoid relying on visual correlations during
    pre-training (Stage 1), then introduces visual conditioning through
    a pose-supervised latent interface (Stage 2). Across four VLA and WAM
    architectures, LIT improves out-of-distribution success rates while
    preserving or improving in-distribution success rates for LIBERO and LIBERO-Plus (center).}
    \label{fig:motivation}
\end{figure*}

Recent robot foundation models, including representative
vision--language--action (VLA) and world--action model (WAM)
architectures, combine pretrained vision--language or video
backbones with action experts, achieving strong in-distribution
manipulation performance~\cite{pi05,molmoact2,fastwam,imagewam}.
Visual representations condition action generation alongside
language and robot-state information~\cite{diffusionpolicy,pi0},
as illustrated in Fig.~\ref{fig:motivation} (left).
Robust generalization under visual distribution shifts, however,
remains a challenge~\cite{liberoplus}.

Under visual conditioning, action experts may learn
\emph{vision--action shortcuts}: reliance on task-irrelevant
visual cues that correlate with demonstrated actions during
training but become unreliable under distribution
shifts~\cite{liberoplus,shortcutlearning}.
Limited visual diversity in robot demonstrations can further
encourage this reliance, undermining generalization to changes
in viewpoint, appearance, or sensing conditions.
Yet visual representations also encode task-relevant spatial
information essential for action generation.
Mitigating these shortcuts therefore requires reducing sensitivity
to task-irrelevant visual variations while preserving
responsiveness to task-relevant changes that require different
actions.

Existing efforts improve generalization through representation
enhancement and stage-wise action pretraining.
Representation-enhanced methods introduce structured spatial
or motion information to ground action generation in
task-relevant geometry~\cite{spatialvla,molmoact,tracevla}.
However, enriching the available information does not directly
constrain how the action expert uses it, leaving room for
vision--action shortcuts.
Stage-wise approaches first learn language-conditioned action
priors without images, then use them to initialize visual policy
training~\cite{la4vla,qwenvla}. However, these approaches lack an
explicit spatial goal for each action chunk during pretraining,
and subsequent training introduces visual conditioning without
explicitly constraining its use. Image-free pretraining alone
therefore leaves the policy susceptible to shortcuts once
visual conditioning is introduced.
These limitations motivate a central question:
\emph{How can action learning and visual conditioning be
structured to mitigate vision--action shortcuts while preserving
the task-relevant spatial information needed for action generation?}

To address this challenge, we propose Latent Interface Training
(LIT), a model-agnostic two-stage strategy that first learns
a spatial-goal-conditioned action prior without images, then
introduces visual conditioning through a pose-supervised latent
interface (Fig.~\ref{fig:motivation}, right).
\textbf{In the first stage}, the action expert is conditioned
on language and robot-state representations only from a frozen
pretrained backbone, together with an encoding of each
demonstrated action chunk's terminal SE(3) end-effector pose.
This establishes a spatial-goal-conditioned action prior
without visual inputs.
\textbf{In the second stage}, learnable latent tokens
cross-attend to visual and semantic backbone representations
and provide layer-wise conditioning to the pretrained action
expert, serving as its only visual conditioning pathway.
A pose-reconstruction objective supervises these tokens to
recover the same terminal pose used in Stage~1, encouraging
them to retain goal-relevant spatial information.
The shared spatial target thus connects action-prior learning
with visual interface learning, while the exclusive interface
constrains visual conditioning to mitigate vision--action
shortcuts.

Our contributions are threefold. First, we introduce
\textbf{Latent Interface Training (LIT)}, a model-agnostic training
strategy for improving generalization in robot foundation
models. LIT combines image-free, spatial-goal-conditioned action
pretraining with a pose-supervised latent interface that serves
as the action expert's only visual conditioning pathway.
Second, across $\pi_{0.5}$~\cite{pi05}, MolmoAct2~\cite{molmoact2}, FAST-WAM~\cite{fastwam}, and ImageWAM~\cite{imagewam},
LIT improves overall LIBERO-Plus~\cite{liberoplus} success by 3.87\%--10.70\%
while preserving or improving average LIBERO~\cite{libero}
success. These results, together with ablations and counterfactual
analyses, support LIT's effectiveness in mitigating vision--action
shortcuts while preserving task-relevant spatial information.
Third, real-world evaluations demonstrate LIT's robustness to
unseen camera configurations, lighting variations, and distractors,
with gains of 13.3--16.7 percentage points in success aggregated
across three manipulation tasks.
\section{Related Work}

We review how visual information conditions action generation in
robot foundation models, followed by representation design for
generalizable control and action pretraining for downstream
policy learning.

\subsection{Visual Conditioning in Robot Foundation Models}

Large-scale cross-embodiment datasets and generalist policies have
expanded the scope of transferable robot control~\cite{rtx,octo}.
Within robot foundation models, visual information enters action
generation through different architectural pathways.
Autoregressive VLAs, such as RT-2~\cite{rt2} and
OpenVLA~\cite{openvla}, predict discretized action tokens from
a shared visual--language context.
Diffusion Policy~\cite{diffusionpolicy} generates action sequences
through conditional denoising, while CogACT~\cite{cogact}
conditions a specialized diffusion action module on VLM
representations.
For flow-based action generation, $\pi_0$~\cite{pi0} and
$\pi_{0.5}$~\cite{pi05} adopt a mixture-of-transformers design,
where the backbone and action expert use separate parameters
and interact through joint attention over their tokens.
MolmoAct2~\cite{molmoact2} instead uses layer-wise cross-attention
to condition its action expert on projected key--value
representations from corresponding backbone layers.
WAMs, including DreamZero~\cite{dreamzero},
FAST-WAM~\cite{fastwam}, and ImageWAM~\cite{imagewam},
incorporate representations learned through video or world
modeling into action generation.

These architectures commonly expose action-generation components
to rich visual representations.
Knowledge insulation~\cite{knowledgeinsulation} improves training
and knowledge transfer by blocking gradients from the action expert
into the backbone, while retaining backbone representations as
conditioning inputs to the action expert.
Learning actions from these representations may encourage reliance
on scene-specific correlations, particularly when training demonstrations
exhibit limited visual diversity. This is consistent with shortcut
learning and causal confusion in imitation
learning~\cite{shortcutlearning,causalconfusion}.
Evaluation studies further document sensitivity to visual shifts
and broader generalization challenges in language-conditioned
manipulation~\cite{liberoplus,simpler,vlabench}.

\subsection{Representations for Generalizable Robot Control}

Prior work improves policy generalization by enriching the
information used for action generation.
R3M~\cite{r3m} and MVP~\cite{mvp} learn transferable visual
features through large-scale video or image pretraining.
Other methods introduce spatial, temporal, or semantic structure.
TraceVLA~\cite{tracevla} augments observations with historical
motion traces, while SpatialVLA~\cite{spatialvla} incorporates
egocentric 3D position information.
ECoT~\cite{ecot} generates textual reasoning about plans,
subtasks, and visually grounded features before predicting actions.
MolmoAct~\cite{molmoact} structures action prediction through
depth tokens and visual reasoning traces, while
CoT-VLA~\cite{cotvla} predicts future images as visual subgoals.
Hierarchical approaches such as HAMSTER~\cite{hamster} and
3D HAMSTER~\cite{hamster3d} guide low-level control with
predicted 2D and 3D end-effector trajectories, respectively.
Crossway Diffusion~\cite{crosswaydiffusion} instead improves
control representations through auxiliary state reconstruction.

Another line of work improves robustness by filtering
task-irrelevant information.
OREO~\cite{oreo} uses object-aware regularization to discourage
reliance on nuisance visual cues, while Selective Visual
Representations~\cite{selectivevisual} learns a task-conditioned
codebook bottleneck to filter visual features.
However, richer representations, explicit guidance, and information
filtering do not by themselves ensure that visual conditioning
retains task-relevant spatial information without exploiting
shortcut-prone cues.
LIT targets this challenge through a pose-supervised latent
interface as the action expert's only visual conditioning pathway.

\subsection{Action Expert Pretraining for Generalizable Policies}

A related line of work pretrains action-generation components to
learn reusable priors for subsequent policy learning.
Qwen-VLA~\cite{qwenvla} pretrains a flow-matching decoder
conditioned on language and embodiment prompts without images,
then introduces visual conditioning.
LA4VLA~\cite{la4vla} learns action priors from atomic instructions
and robot states without images; its sequential LA-to-VLA variant
initializes subsequent visual policy training with these priors.
Action prior learning~\cite{jing2026learning} uses state--action
trajectories without images or language, transferring the learned
prior through decoder reuse and early-stage latent distillation.
APT~\cite{apt} instead pretrains a vision--action expert before
introducing language conditioning to improve generalization
to unseen instructions.

These studies show that action pretraining can facilitate
downstream policy learning and improve generalization in
different settings. However, pretraining alone does not explicitly
constrain how visual information is used during subsequent policy
learning, leaving action generation susceptible to scene-specific
visual correlations.
LIT uses each action chunk's terminal pose to guide image-free
action pretraining, then introduces visual conditioning through
a latent interface supervised to reconstruct the same pose.
This interface serves as the action expert's only visual
conditioning pathway.
\section{Method}

\begin{figure*}[t]
    \centering
    \includegraphics[width=\textwidth]{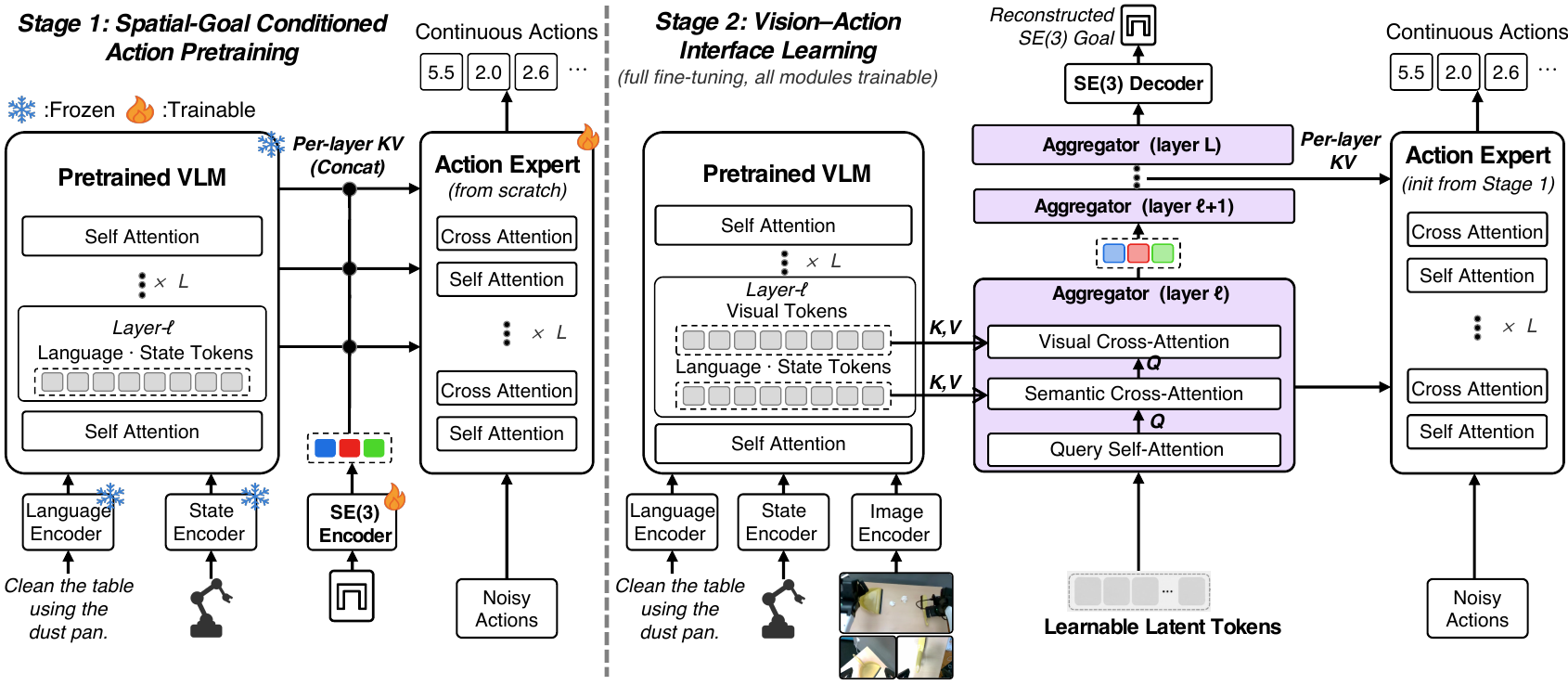}
    \vspace{-10pt}
    \caption{
    \textbf{Overview of Latent Interface Training (LIT).}
    \textbf{Stage 1} learns a spatial-goal-conditioned action prior without
    images by conditioning the action expert on language, robot state,
    and each demonstrated action chunk's terminal SE(3) end-effector pose.
    \textbf{Stage 2} introduces visual conditioning through learnable
    latent tokens that aggregate visual and semantic backbone representations
    and provide layer-wise conditioning to the pretrained action expert.
    The tokens are supervised to reconstruct the same terminal pose,
    linking action-prior learning with visual interface learning.
    Routing all visual conditioning through these spatially supervised
    tokens aims to mitigate vision--action shortcuts while preserving
    task-relevant spatial information.
    }
    \vspace{-5pt}
    \label{fig:framework}
\end{figure*}

Figure~\ref{fig:framework} provides an overview of \textbf{Latent
Interface Training (LIT)}, a model-agnostic two-stage strategy
for reducing vision--action shortcuts while preserving task-relevant
spatial information. LIT first establishes an action prior without
images, then introduces visual conditioning exclusively through a
pose-supervised latent interface. We first formulate the problem,
then describe spatial-goal-conditioned action pretraining
(Sec.~\ref{sec:stage1}) and latent interface learning
(Sec.~\ref{sec:stage2}), followed by framework integration and
inference (Sec.~\ref{sec:integration}).

\subsection{Problem Setup}

We consider robot foundation models that combine a pretrained
backbone with an embodiment-specific action expert.
At timestep $t$, the policy receives a visual observation
$\mathbf{o}_t$, a language instruction $l$, and a robot state
$\mathbf{s}_t$, and predicts an action chunk of horizon $H$:
\begin{equation}
    \mathbf{A}_t
    =
    \left(\mathbf{a}_t,\ldots,\mathbf{a}_{t+H-1}\right).
\end{equation}
For each demonstrated chunk, we use the terminal robot state
as the goal:
\begin{equation}
    \mathbf{g}_t
    =
    \left[
        \mathbf{p}_{t+H};
        \mathbf{r}_{t+H};
        \mathbf{q}_{t+H}
    \right]
    \in \mathbb{R}^{8},
\end{equation}
where $\mathbf{p}\in\mathbb{R}^{3}$ and
$\mathbf{r}\in\mathbb{R}^{3}$ represent the world-frame end-effector
position and axis-angle orientation, and
$\mathbf{q}\in\mathbb{R}^{2}$ contains the gripper joint positions.
This goal serves as a conditioning signal in Stage~1 and a
reconstruction target in Stage~2, and is not required at inference.

In standard architectures, backbone visual representations directly
condition the action expert. LIT instead routes visual conditioning
through a pose-supervised latent interface to an action expert
pretrained without images, as detailed below.

\subsection{Stage 1: Spatial-Goal-Conditioned Action Pretraining}
\label{sec:stage1}

Stage~1 trains the action expert from scratch to generate demonstrated
action chunks conditioned on language, robot state, and the terminal
pose $\mathbf{g}_t$, learning a spatial-goal-conditioned action prior
without images.

The frozen backbone processes only the language instruction $l$
and robot state $\mathbf{s}_t$, producing representations
$\mathbf{H}_{\ell,t}^{\mathrm{sem}}$ at each of the $L$ coupling layers.
A trainable three-layer MLP with GELU activations maps $\mathbf{g}_t$
to goal tokens $\mathbf{G}_t$, which are concatenated with the
backbone representations:
\begin{equation}
\begin{aligned}
    \mathbf{G}_t
    &= E_{\eta}(\mathbf{g}_t), \\
    \mathbf{C}_{\ell,t}
    &= \left[
        \mathbf{H}_{\ell,t}^{\mathrm{sem}};
        \mathbf{G}_t
    \right],
    \qquad \ell=1,\ldots,L.
\end{aligned}
\label{eq:stage1-conditioning}
\end{equation}
Here, $[\cdot;\cdot]$ denotes token concatenation.
$\mathbf{C}_{\ell,t}$ conditions the corresponding action expert
layer through the architecture's native conditioning mechanism.

We retain each framework's native action-generation objective.
For the flow-matching objective used here~\cite{flowmatching},
we sample $\tau\sim\mathcal{U}(0,1)$ and
$\boldsymbol{\epsilon}\sim\mathcal{N}(\mathbf{0},\mathbf{I})$,
and construct the noisy action chunk and target velocity:
\begin{equation}
\begin{aligned}
    \widetilde{\mathbf{A}}_t^{\tau}
    &= (1-\tau)\boldsymbol{\epsilon}+\tau\mathbf{A}_t, \\
    \mathbf{v}_t^{\star}
    &= \mathbf{A}_t-\boldsymbol{\epsilon}.
\end{aligned}
\label{eq:stage1-flow-path}
\end{equation}
Given $\mathbf{C}_{1:L,t}
=\{\mathbf{C}_{\ell,t}\}_{\ell=1}^{L}$,
the action expert predicts the velocity and minimizes:
\begin{equation}
\begin{aligned}
    \widehat{\mathbf{v}}_t^{\tau}
    &= v_{\theta}\!\left(
        \widetilde{\mathbf{A}}_t^{\tau},
        \tau;
        \mathbf{C}_{1:L,t}
    \right), \\
    \mathcal{L}_{\mathrm{prior}}
    &= \mathbb{E}_{\mathbf{A}_t,\tau,\boldsymbol{\epsilon}}
    \left[
        \left\|
            \widehat{\mathbf{v}}_t^{\tau}
            -\mathbf{v}_t^{\star}
        \right\|_2^2
    \right].
\end{aligned}
\label{eq:stage1-objective}
\end{equation}
Only the action expert parameters $\theta$ and SE(3) encoder
parameters $\eta$ are updated; the backbone and its modality
encoders remain frozen. The learned action expert parameters
initialize Stage~2.

\subsection{Stage 2: Vision--Action Interface Learning}
\label{sec:stage2}

Stage~2 initializes the action expert from Stage~1 and introduces
visual conditioning exclusively through a pose-supervised latent
interface.

Let $\mathbf{Z}^{0}\in\mathbb{R}^{K\times d}$ denote learnable
latent tokens shared across inputs, with $K=100$ and token dimension
$d$. For each policy input, the interface starts from
$\mathbf{Z}_{0,t}=\mathbf{Z}^{0}$.
At coupling layer $\ell$, the backbone provides language and state
representations $\mathbf{H}_{\ell,t}^{\mathrm{sem}}$ and visual
representations $\mathbf{H}_{\ell,t}^{\mathrm{vis}}$.
The latent tokens are updated through self-attention, semantic
cross-attention, and visual cross-attention:
\begin{equation}
\begin{aligned}
    \overline{\mathbf{Z}}_{\ell,t}
    &=
    \mathbf{Z}_{\ell-1,t}
    +
    \operatorname{SA}_{q(\ell)}
    \left(\mathbf{Z}_{\ell-1,t}\right), \\
    \widetilde{\mathbf{Z}}_{\ell,t}
    &=
    \overline{\mathbf{Z}}_{\ell,t}
    +
    \operatorname{CA}_{q(\ell)}^{\mathrm{sem}}
    \left(
        \overline{\mathbf{Z}}_{\ell,t};
        \mathbf{H}_{\ell,t}^{\mathrm{sem}}
    \right), \\
    \mathbf{Z}_{\ell,t}
    &=
    \widetilde{\mathbf{Z}}_{\ell,t}
    +
    \operatorname{CA}_{q(\ell)}^{\mathrm{vis}}
    \left(
        \widetilde{\mathbf{Z}}_{\ell,t};
        \mathbf{H}_{\ell,t}^{\mathrm{vis}}
    \right),
    \quad \ell=1,\ldots,L.
\end{aligned}
\label{eq:latent-token-update}
\end{equation}
Latent tokens provide the queries in each cross-attention operation,
while backbone representations provide the keys and values.
For parameter efficiency, every $m$ consecutive coupling layers
share interface attention parameters, indexed by
$q(\ell)=\lceil\ell/m\rceil$, while accessing their respective
backbone representations.

The updated tokens condition the corresponding action expert layers
through the architecture's native conditioning mechanism.
Using Eq.~\ref{eq:stage1-flow-path}, the action expert predicts
\begin{equation}
    \widehat{\mathbf{v}}_t^\tau
    =
    v_\theta\!\left(
        \widetilde{\mathbf{A}}_t^\tau,
        \tau;
        \mathbf{Z}_{1:L,t}
    \right),
\label{eq:stage2-velocity}
\end{equation}
where $\mathbf{Z}_{1:L,t}
=\{\mathbf{Z}_{\ell,t}\}_{\ell=1}^{L}$.
The action loss $\mathcal{L}_{\mathrm{act}}$ follows
Eq.~\ref{eq:stage1-objective}, replacing
$\mathbf{C}_{1:L,t}$ with $\mathbf{Z}_{1:L,t}$.

An MLP decoder reconstructs the same terminal goal state
$\mathbf{g}_t$ used to condition Stage~1 from the final latent tokens:
\begin{equation}
\begin{aligned}
    \widehat{\mathbf{g}}_t
    &= D_\omega(\mathbf{Z}_{L,t}), \\
    \mathcal{L}_{\mathrm{pose}}
    &= 
    \left\|\widehat{\mathbf{g}}_t-\mathbf{g}_t\right\|_2^2, \\
    \mathcal{L}_{\mathrm{stage2}}
    &= \mathcal{L}_{\mathrm{act}}
    + \lambda_{\mathrm{pose}}\mathcal{L}_{\mathrm{pose}}.
\end{aligned}
\label{eq:stage2-objective}
\end{equation}
We set $\lambda_{\mathrm{pose}}=0.3$. The reconstruction loss is
computed in the preprocessed state space and averaged over valid
targets, encouraging the interface to retain goal-relevant information.
Stage~2 jointly optimizes the backbone and its modality encoders,
the action expert, the latent tokens and interface attention modules,
and the MLP decoder.

\subsection{Framework Integration and Inference}
\label{sec:integration}

LIT introduces a latent interface between a pretrained backbone
(VLM or video model) and an action expert, making it applicable to
VLA and WAM architectures with this structure. The interface provides
layer-wise conditioning through each architecture's native mechanism,
while retaining the backbone and action expert architectures,
action representations, prediction horizons, and native
action-generation objectives.

At inference, the latent interface remains active, while the
Stage~1 pose encoder and Stage~2 pose decoder are omitted.
The policy requires only visual observations, language, and robot
state, and follows each framework's native action sampling and
execution procedure.
\section{Experiments}
\label{sec:experiments}

Our experimental evaluation comprises a broad suite of studies designed to assess the effectiveness, robustness, and generality of LIT across different robot foundation model architectures and deployment settings. Specifically, we investigate LIT along five main dimensions: (i) we examine whether LIT can preserve or improve in-distribution task performance across heterogeneous VLA and WAM architectures; (ii) we evaluate whether LIT improves zero-shot generalization under a diverse set of task-preserving distribution shifts, including changes in camera viewpoints, sensor noise, lighting, background textures, robot initial states, object layouts, and language instructions; (iii) we assess whether these generalization gains transfer to real-world manipulation under changes in lighting, camera configuration, and task-irrelevant distractors; (iv) we analyze whether LIT reduces reliance on spurious vision--action shortcuts by maintaining task-relevant visual attention and responding appropriately to visual and goal interventions; and (v) we study the learning dynamics and key design components of LIT, including Stage~1 spatial-goal-conditioned action-prior learning, pose supervision, and restricting visual conditioning to the latent interface, to determine which factors are responsible for its generalization gains.

\subsection{Experimental Setup}

\noindent\textbf{Benchmarks and Evaluation Protocol}
We evaluate on LIBERO and LIBERO-Plus using task success rate as
the metric. We evaluate 40 tasks across the LIBERO-Spatial,
LIBERO-Object, LIBERO-Goal, and LIBERO-Long suites~\cite{libero}.
We conduct 50 rollouts per task, totaling 2,000 episodes, to evaluate
performance on the original task distribution.
LIBERO-Plus evaluates zero-shot generalization under seven
task-preserving perturbations: camera viewpoints, sensor noise,
robot initial states, language instructions, object layout,
lighting conditions, and background textures~\cite{liberoplus}.
We evaluate all 10,030 perturbed instances using one rollout per
instance with a fixed evaluation seed and report success rates
for each perturbation dimension. Overall denotes the unweighted
mean over the seven perturbation dimensions.
All models are trained only on the original LIBERO demonstrations
and evaluated on LIBERO-Plus without adaptation.

\noindent\textbf{Architectures and Baselines.}
To evaluate LIT across heterogeneous robot foundation model designs,
we integrate it into two VLA architectures,
$\pi_{0.5}$~\cite{pi05} and MolmoAct2~\cite{molmoact2},
and two WAM architectures, FAST-WAM~\cite{fastwam} and
ImageWAM~\cite{imagewam}. These architectures cover two mechanisms
for coupling VLMs with action experts and two different uses of
future visual modeling:

\begin{itemize}
\item $\pi_{0.5}$: a VLA that couples separate VLM and action expert
branches through shared self-attention in a Mixture-of-Transformers
architecture.
\item MolmoAct2: a VLA whose action expert cross-attends to the
corresponding per-layer KV caches of the VLM.
\item FAST-WAM: a WAM that uses future-video prediction during
training but performs action-only inference without future prediction.
\item ImageWAM: a WAM that retains image-editing denoising at inference
and conditions its action expert on the resulting KV caches without
decoding the target image.
\end{itemize}
Despite these architectural differences, all four baselines condition
action generation on rich visual or world-model representations through
their native interaction mechanisms. LIT routes this conditioning
through the pose-supervised latent interface while retaining each
framework's backbone and action expert architectures, action
representation, and action-generation objective.

\noindent\textbf{Training Details.}
For each architecture, the baseline and LIT start from the same
pretrained backbone and randomly initialized action experts.
\textbf{We do not fine-tune pretrained VLA or WAM policy checkpoints},
avoiding vision--action dependencies inherited from policy pretraining.
Our baseline results are therefore not directly comparable to published
results obtained by fine-tuning such checkpoints.
Each baseline--LIT pair uses identical training data and matched shared
settings, following the architecture's native configurations.
Latent interface dimensions are adapted to the backbone and action
expert to match the native conditioning mechanism.
Each baseline follows its architecture's native training budget.
LIT allocates the same total number of optimization steps across
its two stages, initializing the Stage~2 action expert from Stage~1.

\begin{table}[t]
    \centering
    \caption{In-distribution success rates (\%) on LIBERO.
    Better result in each architecture in bold.}
    \label{tab:libero}

    \small
    \setlength{\tabcolsep}{2.2pt}
    \renewcommand{\arraystretch}{1.18}

    % Remove gaps where horizontal and vertical rules meet.
    \setlength{\aboverulesep}{0pt}
    \setlength{\belowrulesep}{0pt}
    \setlength{\abovetopsep}{0pt}
    \setlength{\belowbottomsep}{0pt}

    % Consistent rule thicknesses.
    \setlength{\heavyrulewidth}{0.8pt}
    \setlength{\lightrulewidth}{0.4pt}

    \begin{tabular*}{\columnwidth}{
        @{\extracolsep{\fill}}l|l|rrrrr@{}
    }
        \toprule
        \textbf{Models}
        & \textbf{Variant}
        & \multicolumn{1}{c}{\textbf{Spatial}}
        & \multicolumn{1}{c}{\textbf{Object}}
        & \multicolumn{1}{c}{\textbf{Goal}}
        & \multicolumn{1}{c}{\textbf{Long}}
        & \multicolumn{1}{c}{\textbf{Avg.}} \\
        \midrule

        \multirow{2}{*}{$\pi_{0.5}$}
        & Baseline
        & 88.60
        & 93.40
        & 89.20
        & 79.80
        & 87.75 \\
        & \textbf{LIT}
        & \textbf{90.20}
        & \textbf{98.80}
        & \textbf{93.40}
        & \textbf{84.80}
        & \textbf{91.80} \\
        \midrule

        \multirow{2}{*}{MolmoAct2}
        & Baseline
        & 93.00
        & \textbf{97.80}
        & \textbf{95.40}
        & 87.80
        & 93.50 \\
        & \textbf{LIT}
        & \textbf{94.60}
        & 96.20
        & 95.20
        & \textbf{90.40}
        & \textbf{94.10} \\
        \midrule

        \multirow{2}{*}{FAST-WAM}
        & Baseline
        & 98.20
        & \textbf{100.00}
        & 97.00
        & 95.20
        & 97.60 \\
        & \textbf{LIT}
        & \textbf{98.80}
        & 99.80
        & \textbf{98.40}
        & \textbf{95.40}
        & \textbf{98.10} \\
        \midrule

        \multirow{2}{*}{ImageWAM}
        & Baseline
        & 98.40
        & \textbf{100.00}
        & 97.60
        & \textbf{96.40}
        & 98.10 \\
        & \textbf{LIT}
        & \textbf{99.60}
        & 99.20
        & \textbf{99.20}
        & 95.60
        & \textbf{98.40} \\
        \bottomrule
    \end{tabular*}
    \vspace{-15pt}
\end{table}

\subsection{Performance Preservation Across Tasks}

To determine whether routing visual conditioning through the
pose-supervised latent interface preserves action-generation performance,
we compare LIT with architecture-matched baselines on the four LIBERO
suites. As shown in Table~\ref{tab:libero}, LIT preserves or improves
the average success rate across all four architectures: from 87.75\%
to 91.80\% for $\pi_{0.5}$, from 93.50\% to 94.10\% for MolmoAct2,
from 97.60\% to 98.10\% for FAST-WAM, and from 98.10\% to 98.40\%
for ImageWAM. Despite small variations on individual suites, no
architecture exhibits an average performance degradation, supporting
the effectiveness of the latent interface for in-distribution action
generation across heterogeneous VLA and WAM architectures.

\begin{table*}[t]
    \centering
    \caption{Zero-shot success rates (\%) on LIBERO-Plus.
    $\Delta$: change from the paired baseline.
    Overall: mean over the seven perturbations.
    Better result in each architecture in bold.}
    \label{tab:libero_plus}
    \footnotesize
    \setlength{\tabcolsep}{2.5pt}
    \renewcommand{\arraystretch}{1.15}
    \begin{tabularx}{\textwidth}{
        @{}l
        @{\hspace{8pt}}
        YYZ
        @{\hspace{10pt}}
        YYZ
        @{\hspace{10pt}}
        YYZ
        @{\hspace{10pt}}
        YYZ
        @{}
    }
        \toprule
        \multirow{2}{*}{\textbf{Perturbation}}
        & \multicolumn{3}{c}{$\pi_{0.5}$ (VLA)}
        & \multicolumn{3}{c}{MolmoAct2 (VLA)}
        & \multicolumn{3}{c}{FAST-WAM (WAM)}
        & \multicolumn{3}{c}{ImageWAM (WAM)} \\
        \cmidrule(lr){2-4}
        \cmidrule(lr){5-7}
        \cmidrule(lr){8-10}
        \cmidrule(lr){11-13}
        & Base & LIT & $\Delta$
        & Base & LIT & $\Delta$
        & Base & LIT & $\Delta$
        & Base & LIT & $\Delta$ \\
        \midrule
        Camera Viewpoints
        & 58.29 & \textbf{80.30} & \textcolor{blue}{+22.01}
        & 39.40 & \textbf{48.41} & \textcolor{blue}{+9.01}
        & 16.40 & \textbf{43.83} & \textcolor{blue}{+27.43}
        & 82.16 & \textbf{84.55} & \textcolor{blue}{+2.39} \\
        Sensor Noise
        & 79.89 & \textbf{91.94} & \textcolor{blue}{+12.05}
        & 49.03 & \textbf{70.77} & \textcolor{blue}{+21.74}
        & 37.70 & \textbf{56.89} & \textcolor{blue}{+19.19}
        & 96.54 & \textbf{97.57} & \textcolor{blue}{+1.03} \\
        Lighting Conditions
        & 82.40 & \textbf{90.11} & \textcolor{blue}{+7.71}
        & 84.86 & \textbf{85.64} & \textcolor{blue}{+0.78}
        & 78.20 & \textbf{83.89} & \textcolor{blue}{+5.69}
        & 97.65 & \textbf{97.99} & \textcolor{blue}{+0.34} \\
        Background Textures
        & 85.32 & \textbf{86.99} & \textcolor{blue}{+1.67}
        & 89.78 & \textbf{94.42} & \textcolor{blue}{+4.64}
        & 53.70 & \textbf{57.22} & \textcolor{blue}{+3.52}
        & 88.26 & \textbf{94.24} & \textcolor{blue}{+5.98} \\
        Robot Initial States
        & 60.71 & \textbf{63.71} & \textcolor{blue}{+3.00}
        & 50.71 & \textbf{59.03} & \textcolor{blue}{+8.32}
        & 44.50 & \textbf{48.86} & \textcolor{blue}{+4.36}
        & 47.81 & \textbf{60.19} & \textcolor{blue}{+12.38} \\
        Object Layout
        & 64.00 & \textbf{73.11} & \textcolor{blue}{+9.11}
        & 55.90 & \textbf{71.61} & \textcolor{blue}{+15.71}
        & 60.70 & \textbf{64.17} & \textcolor{blue}{+3.47}
        & 77.72 & \textbf{83.67} & \textcolor{blue}{+5.95} \\
        Language Instructions
        & 52.15 & \textbf{71.50} & \textcolor{blue}{+19.35}
        & \textbf{75.69} & 73.58 & \textcolor{red}{-2.11}
        & 68.90 & \textbf{69.55} & \textcolor{blue}{+0.65}
        & \textbf{90.97} & 90.05 & \textcolor{red}{-0.92} \\
        \midrule
        \textbf{Overall}
        & 68.97 & \textbf{79.67}
        & \textcolor{blue}{\textbf{+10.70}}
        & 63.62 & \textbf{71.92}
        & \textcolor{blue}{\textbf{+8.30}}
        & 51.44 & \textbf{60.63}
        & \textcolor{blue}{\textbf{+9.19}}
        & 83.02 & \textbf{86.89}
        & \textcolor{blue}{\textbf{+3.87}} \\
        \bottomrule
    \end{tabularx}
\end{table*}

\subsection{Generalization Across VLA and WAM Architectures}
\label{sec:generalization}

To evaluate whether LIT improves zero-shot generalization across
different robot foundation model families, we compare it with
architecture-matched baselines on LIBERO-Plus. As shown in
Table~\ref{tab:libero_plus}, LIT consistently improves the overall
success rate across all evaluated VLA and WAM architectures.

\begin{figure*}[t]
    \centering
    \includegraphics[width=\textwidth]{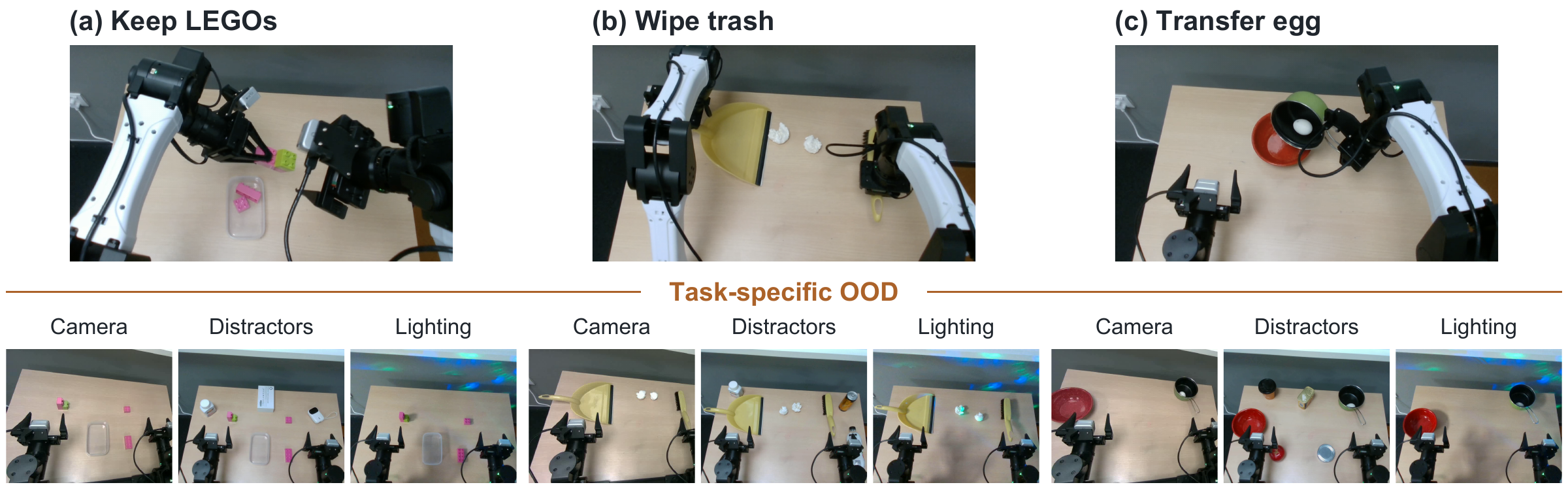}
    \vspace{-15pt}
    \caption{Real-robot evaluation environments under ID and OOD visual conditions. Rows correspond to the three manipulation tasks, while columns show the in-distribution setting and the three evaluated visual shifts: Lighting OOD, Camera OOD, and Distractors OOD. Camera OOD uses only the top-camera
view.}
    \label{fig:real_robot_ood}
\end{figure*}

We first examine the two VLA architectures. LIT improves the overall success rate of MolmoAct2 from 63.62\% to 71.92\%, with substantial gains under Sensor Noise (+21.74 points) and Object Layout (+15.71 points). A similar trend is observed for $\pi_{0.5}$, whose overall success rate increases from 68.97\% to 79.67\%, with gains of 22.01 points under Camera Viewpoints, 12.05 points under Sensor Noise, and 9.11 points under Object Layout. The improvements under viewpoint and sensing perturbations suggest that LIT reduces reliance on nuisance visual correlations, while the gains under object layout demonstrate that it preserves the spatial information necessary for adaptive action generation. Notably, despite relying on different interaction mechanisms, with layerwise cross attention in MolmoAct2 and shared self attention in $\pi_{0.5}$, both models consistently benefit from LIT. These results demonstrate that LIT generalizes across distinct VLA architectures and visual action interaction mechanisms.

We next examine whether LIT extends beyond VLA architectures to world-action models. LIT improves FAST-WAM from 51.44\% to 60.63\% overall, with gains of 27.43 points under Camera Viewpoints and 19.19 under Sensor Noise. ImageWAM improves from 83.02\% to 86.89\%, including gains of 12.38 points under Robot Initial States and 5.95 under Object Layout. FAST-WAM uses future prediction only during training, whereas ImageWAM retains future-image modeling at inference. Improvements in both settings support LIT's applicability across distinct world-action modeling pipelines. Overall, LIT improves 26 of the 28 architecture--perturbation
comparisons, with no decrease exceeding 2.11 percentage points,
supporting its broad effectiveness across the evaluated VLA
and WAM architectures.

\subsection{Real-world Evaluations}
\label{sec:real_world}

To evaluate whether the generalization improvements observed in simulation
transfer to physical manipulation, we compare LIT with the
architecture-matched MolmoAct2 baseline on three real-robot tasks.
In \texttt{Keep LEGOs}, the robot must detach all blocks and put them into a box.
In \texttt{Wipe trash}, the robot must clean the table using a dustpan.
In \texttt{Transfer egg}, the robot must transfer an egg from a pan into a bowl.
For each method, we jointly train a single multi-task policy on 300
demonstrations, with 100 demonstrations per task, and evaluate the resulting
policy separately on each task.

\begin{figure}[t]
    \centering
    \includegraphics[width=\linewidth]{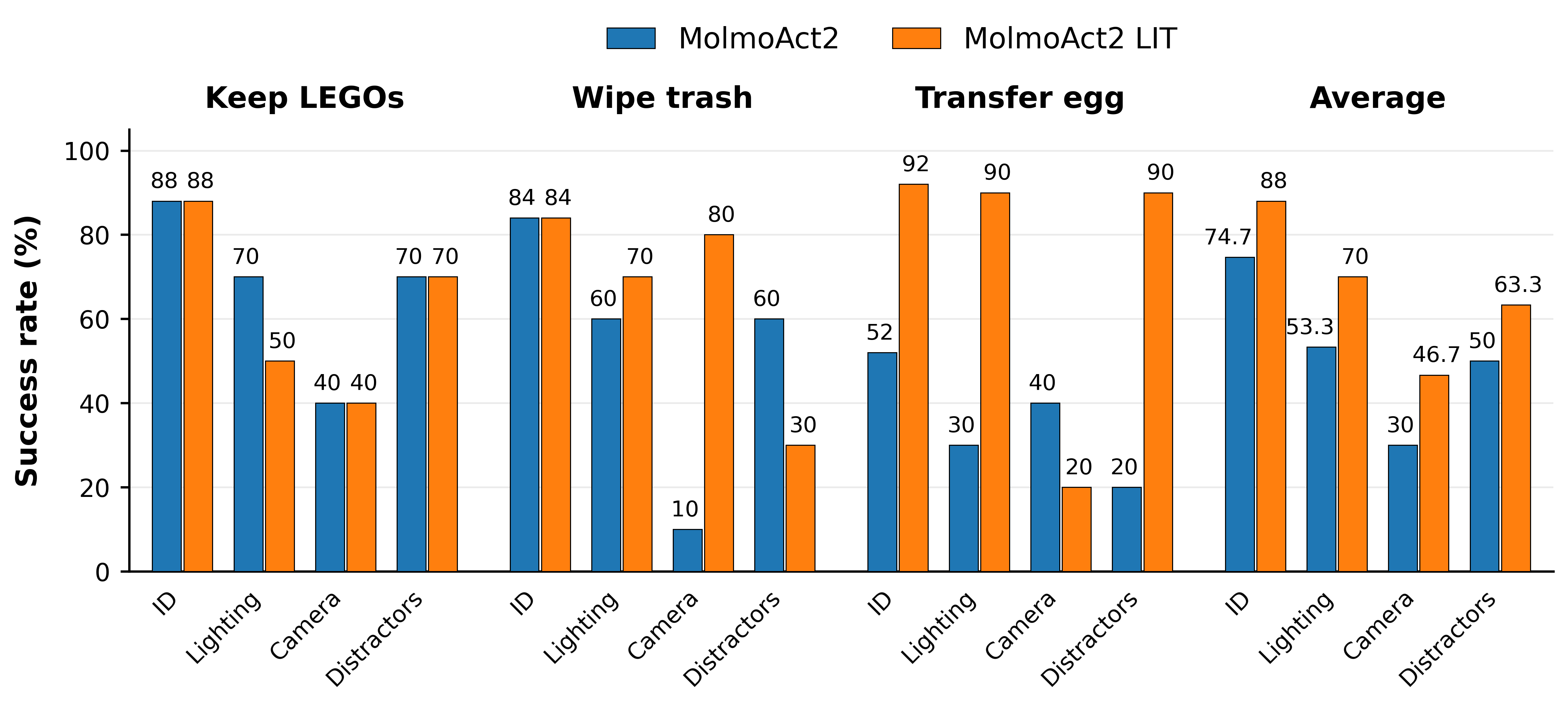}
    \caption{Real-robot task success under in-distribution (ID) and
        out-of-distribution (OOD) visual conditions. Camera OOD uses only
        the top-camera view. Bars report success rates (\%).}
    \label{fig:real_robot_success}
\end{figure}

We evaluate each policy under an in-distribution (ID) setting and three
out-of-distribution (OOD) visual conditions, as shown in
Fig.~\ref{fig:real_robot_ood}. In \textbf{Lighting OOD},
the scene illumination is changed from that observed during training.
In \textbf{Camera OOD}, the policy is evaluated by changing the camera
configuration (top-camera only) from the training setting.
In \textbf{Distractors OOD}, additional task-irrelevant objects are
introduced into the scene while leaving the task unchanged.
These perturbations test robustness to changes in visual appearance,
camera configuration, and scene content, respectively.
Each policy is evaluated for 25 rollouts per task under ID conditions
(75 rollouts total) and 10 rollouts per task under each OOD condition
(90 OOD rollouts total).
Fig.~\ref{fig:real_robot_success} reports success rates.

\begin{figure*}[t]
\centering
\includegraphics[width=\textwidth]{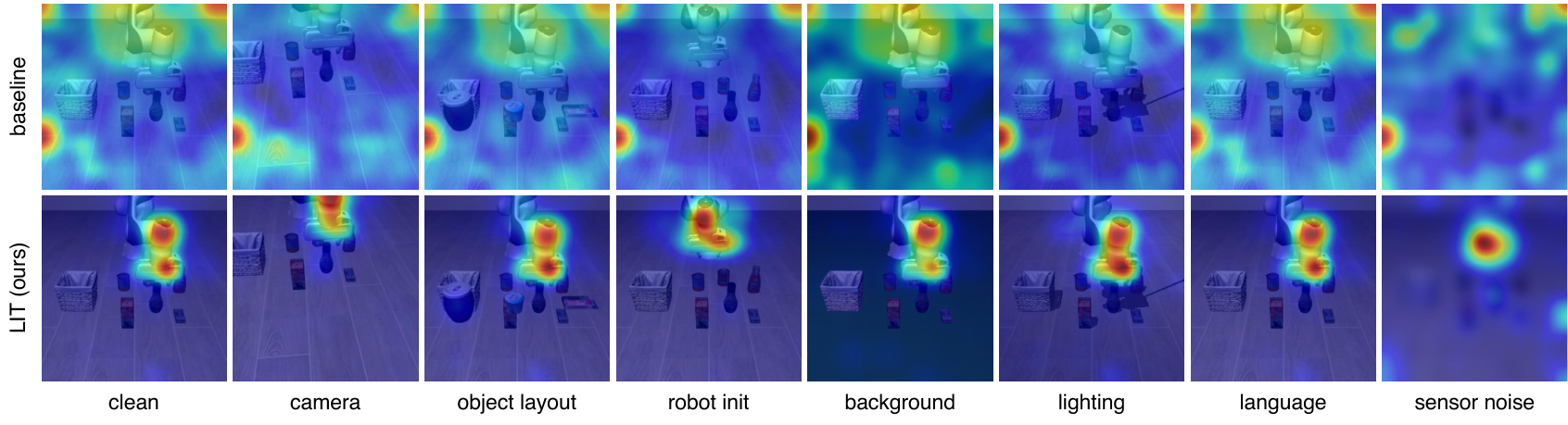}
\vspace{-20pt}
\caption{Effective action-to-image attention under distribution
shifts. Baseline attention varies substantially across
perturbations, whereas LIT remains consistently aligned with
task-relevant robot--object regions.}
\label{fig:attention}
\end{figure*}

\begin{figure}[t]
\centering
\includegraphics[width=\columnwidth]{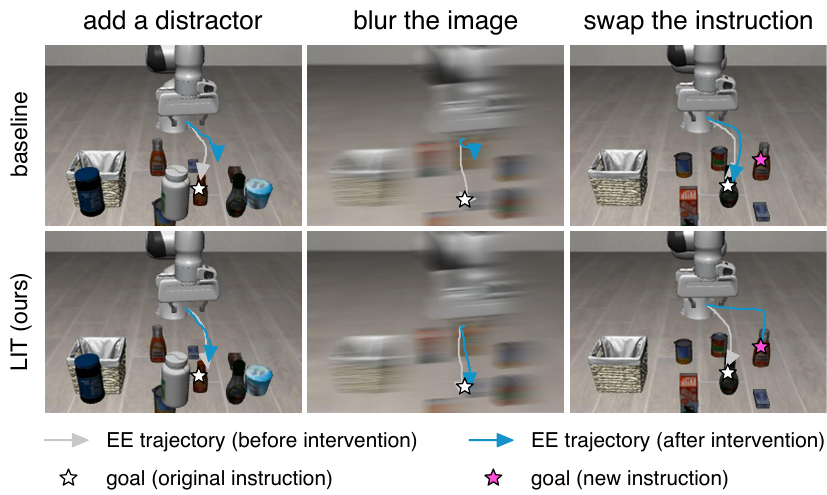}
\vspace{-20pt}
\caption{Counterfactual trajectory analysis. LIT remains stable
under task-preserving visual interventions and redirects its
trajectory when the specified goal changes.}
\label{fig:intervention}
\end{figure}

The real-robot results broadly support the generalization trends observed
in simulation. Aggregated across the three tasks, LIT improves ID success
from 74.7\% to 88.0\%. More importantly, the improvement persists under
all three OOD conditions. LIT improves success from 53.3\% to 70.0\%
under Lighting OOD, from 30.0\% to 46.7\% under Camera OOD, and from
50.0\% to 63.3\% under Distractors OOD.

The gains are particularly pronounced on \texttt{Transfer egg}, where LIT improves
ID success from 52.0\% to 92.0\%. Under Lighting OOD and Distractors
OOD, success increases from 30.0\% to 90.0\% and from 20.0\% to 90.0\%,
respectively. On \texttt{Wipe trash}, LIT matches the baseline under ID conditions
while substantially improving Camera OOD performance from 10.0\% to
80.0\%. These results indicate that the generalization benefits of the
latent interface extend beyond simulated perturbations to physical
changes in the robot's visual observations.

While the magnitude of the improvement varies across tasks and
perturbation types, the overall trend favors LIT. Aggregated across the
three tasks, LIT achieves higher success under ID conditions and under
each of the three OOD conditions. Together with the simulation results,
these findings provide complementary evidence that the latent interface
improves robustness to visual distribution shifts in physical
manipulation.

% Place this figure in the real-robot evaluation subsection.

\subsection{Analysis and Ablation Studies}
Having established LIT's generalization improvements in
Table~\ref{tab:libero_plus}, we investigate the behavior,
learning dynamics, and design choices underlying these gains.
Unless otherwise specified, all analyses and ablations use
MolmoAct2. We first examine visual attention and responses to
controlled interventions, then analyze learning dynamics across
the two training stages, and finally evaluate the contributions
of LIT's components through six ablations.

\subsubsection{Behavioral Analysis}

\noindent\textbf{Stability of action-to-image attention.}
We examine whether LIT maintains attention to task-relevant
regions under distribution shifts by visualizing attention maps
for clean and perturbed observations.
For the baseline, we use direct action-to-visual attention.
For LIT, we compose attention from the action expert to the
latent tokens with attention from the latent tokens to visual
representations.
Figure~\ref{fig:attention} compares the baseline (top row) and
LIT (bottom row). Baseline attention shifts across perturbations,
whereas LIT's attention remains more concentrated on task-relevant
robot--object regions. These qualitative observations suggest
more stable visual attention during action generation.

\noindent\textbf{Responses to visual and goal interventions.}
We probe reliance on vision--action shortcuts through two types
of counterfactual interventions (Fig.~\ref{fig:intervention}).
For \emph{task-preserving visual interventions}, we add a
distractor or blur the image while keeping the instruction,
spatial goal, and robot state fixed (left and middle columns).
The baseline trajectories deviate substantially from their clean
counterparts, whereas LIT's remain close.
For the \emph{goal-changing intervention}, we change the
instructed goal while keeping the visual scene and robot state
fixed (right column). The baseline continues toward the original
goal, consistent with reliance on unchanged visual cues,
whereas LIT redirects toward the new goal.
Together, LIT's robustness to task-irrelevant visual changes
and responsiveness to goal changes provide evidence consistent
with reduced reliance on vision--action shortcuts.

\subsubsection{Learning Dynamics}

\begin{figure*}[t]
    \centering
    \includegraphics[width=\textwidth]{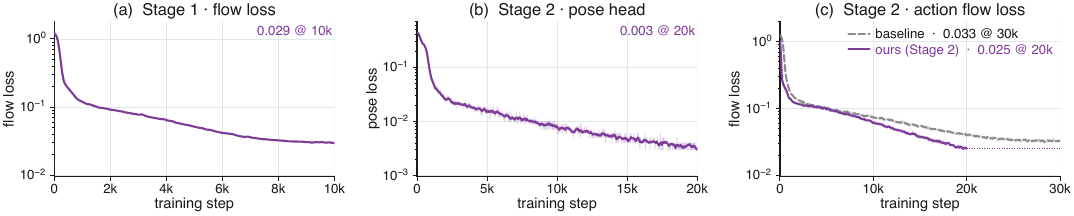}
    \caption{\textbf{Training dynamics on MolmoAct2.}
    (a) Stage~1 action flow-matching loss.
    (b) Stage~2 pose-reconstruction loss.
    (c) Action flow-matching loss during visual policy training.
    LIT uses 10K Stage~1 and 20K Stage~2 steps, matching the
    baseline's 30K-step total budget.
    LIT's Stage~2 action loss reaches a lower value than the
    baseline's final loss.
    Panel (c) counts visual training steps; the 0.63$\times$
    ratio compares losses at 20K visual training steps.}
    \label{fig:training_loss}
\end{figure*}

We examine whether the training dynamics align with the intended
roles of action-prior learning and visual-interface learning
(Fig.~\ref{fig:training_loss}).
In Stage~1, the action loss decreases rapidly to 0.029 after
10K steps, indicating that the action expert learns
goal-directed action generation without images.
In Stage~2, the pose-reconstruction loss reaches 0.003 after
20K steps, showing that the demonstrated spatial goals are
recoverable from the learned interface on training data.
LIT also achieves a lower action loss in 20K visual training
steps than the baseline reaches in 30K steps, with both methods
using the same total training budget.
These dynamics support the intended learning roles of the two
stages and complement the behavioral analyses and generalization
results.

\subsubsection{Component Ablations}

We evaluate six ablations in Table~\ref{tab:ablation}.
We first examine the contributions of action pretraining,
pose supervision, and the restricted visual pathway by removing
individual components from LIT. We then test whether latent-token
aggregation, staged training, or pose supervision alone can
account for its gains.
\begin{table*}[t]
    \centering
    \caption{\textbf{Ablation study of LIT on MolmoAct2.}
    Task success rates (\%) on LIBERO (ID) and LIBERO-Plus (OOD).
    Overall denotes aggregate OOD success. Higher is better.
    Component ablations remove individual elements from LIT;
    alternative designs test simpler explanations for its gains.
    The best and second-best results in each column are shown in
    \textbf{bold} and \underline{underlined}, respectively.}
    \label{tab:ablation}
    \setlength{\tabcolsep}{4pt}
    \renewcommand{\arraystretch}{1.15}
    \resizebox{\textwidth}{!}{%
    \begin{tabular}{@{}l*{9}{c}@{}}
        \toprule
        & ID & \multicolumn{8}{c}{OOD: LIBERO-Plus} \\
        \cmidrule(lr){2-2}
        \cmidrule(lr){3-10}
        Variant
        & \shortstack{LIBERO\\Avg.}
        & \shortstack{Camera\\Viewpoints}
        & \shortstack{Sensor\\Noise}
        & \shortstack{Lighting\\Conditions}
        & \shortstack{Background\\Textures}
        & \shortstack{Robot Initial\\States}
        & \shortstack{Object\\Layout}
        & \shortstack{Language\\Instructions}
        & Overall \\
        \midrule
        MolmoAct2
        & 93.50 & 39.40 & 49.03 & 84.86 & 89.78
        & 50.71 & 55.90 & \textbf{75.69} & 63.62 \\
        \midrule
        \multicolumn{10}{@{}l}{\textit{Component ablations}} \\
        LIT w/o Stage 1
        & 93.70 & \underline{46.15} & 54.65 & 87.13 & 86.34
        & \textbf{63.10} & 66.43 & \underline{73.80} & 68.23 \\
        LIT w/o pose supervision
        & 93.50 & 42.78 & \underline{60.02} & \textbf{91.07} & 88.38
        & 57.81 & \underline{68.85} & 73.11 & \underline{68.86} \\
        LIT w/ direct visual access
        & \textbf{94.25} & 42.96 & 55.28 & 88.18 & 89.22
        & \underline{62.49} & 63.74 & 72.33 & 67.74 \\
        \midrule
        \multicolumn{10}{@{}l}{\textit{Alternative designs}} \\
        LIT w/o Stage 1 \& pose supervision
        & 93.45 & 39.84 & 49.34 & 87.22 & 89.63
        & 53.00 & 68.20 & 72.67 & 65.70 \\
        LA4VLA-inspired staged training~\cite{la4vla}
        & 93.75 & 43.28 & 52.09 & 85.17 & 88.08
        & 55.16 & 60.98 & 73.46 & 65.46 \\
        Baseline w/ pose supervision
        & 93.20 & 40.32 & 51.30 & \underline{89.00} & \underline{90.13}
        & 55.45 & 58.37 & 73.56 & 65.45 \\
        \midrule
        \textbf{LIT}
        & \underline{94.10} & \textbf{48.41} & \textbf{70.77}
        & 85.64 & \textbf{94.42}
        & 59.03 & \textbf{71.61} & 73.58 & \textbf{71.92} \\
        \bottomrule
    \end{tabular}%
    }
\end{table*}

\noindent\textbf{Contribution of the action prior.}
We skip Stage~1 and randomly initialize the action expert,
while retaining the Stage~2 latent interface and pose supervision
(\emph{LIT w/o Stage 1}).
OOD success decreases from 71.92\% to 68.23\%.
This 3.69-percentage-point decrease supports the benefit of
spatial-goal-conditioned action pretraining for subsequent
visual policy learning.

\noindent\textbf{Contribution of pose supervision.}
We remove the Stage~2 pose-reconstruction loss while retaining
all other components (\emph{LIT w/o pose supervision}).
OOD success decreases from 71.92\% to 68.86\%, a reduction of
3.06 percentage points.
The largest decreases occur under Sensor Noise
(from 70.77\% to 60.02\%, 10.75 percentage points) and
Background Textures (from 94.42\% to 88.38\%, 6.04 points).
These results are consistent with spatial-goal supervision
improving robustness to task-irrelevant visual perturbations.

\noindent\textbf{Contribution of the restricted visual pathway.}
We allow the action expert to directly access backbone visual
representations while retaining all LIT components
(\emph{LIT w/ direct visual access}).
Although LIBERO success remains comparable to LIT
(94.25\% vs.\ 94.10\%), LIBERO-Plus success decreases from
71.92\% to 67.74\%.
This 4.18-percentage-point decrease supports the importance
of routing visual information exclusively through the
pose-supervised latent interface for generalization.

\noindent\textbf{Latent-token aggregation alone.}
To assess whether latent-token aggregation accounts for LIT's
gains, we jointly remove Stage~1 and pose supervision
(\emph{LIT w/o Stage 1 \& pose supervision}).
Learnable tokens still aggregate backbone representations to
condition the action expert, as in query-based interfaces such
as VLA-Adapter~\cite{wang2026vlaadapter}, but are trained solely
through the action objective.
OOD success reaches 65.70\%, only 2.08 percentage points above
the baseline and 6.22 points below LIT.
This result supports combining the latent interface with
action pretraining and spatial-goal supervision.

\noindent\textbf{Staged training alone.}
To assess whether two-stage training accounts for LIT's gains,
we evaluate an LA4VLA-inspired baseline~\cite{la4vla}
(\emph{LA4VLA-inspired staged training}).
It first predicts actions from language and robot state without
images, then conditions the action expert on backbone visual
representations.
OOD success increases from 63.62\% for MolmoAct2 to 65.46\%,
but remains 6.46 percentage points below LIT.
Thus, the tested staged-training alternative does not account
for LIT's full gains.

\noindent\textbf{Pose supervision alone.}
We add the same pose-reconstruction objective to the baseline
while retaining its original visual conditioning architecture
(\emph{Baseline w/ pose supervision}).
An MLP decoder reconstructs the terminal goal state from the
baseline's final-layer backbone features.
OOD success increases from 63.62\% to 65.45\%, but remains
6.47 percentage points below LIT.
This modest improvement suggests that pose supervision alone
does not explain LIT's gains, supporting its integration with
the action prior and restricted latent interface.

\section{Limitations and Conclusion}
While our experiments demonstrate improved robustness to visual
variations, evaluation at larger real-world scales remains open.
Future work will explore scaling LIT to larger and more diverse
robot demonstration datasets and extending its evaluation to
a broader range of tasks and environments.

We presented Latent Interface Training (LIT), a framework-agnostic
two-stage strategy for mitigating vision--action shortcuts in
robot foundation models. LIT combines image-free,
spatial-goal-conditioned action pretraining with a pose-supervised
latent interface as the action expert's only visual conditioning pathway.
Across four VLA and WAM architectures, LIT preserves
in-distribution performance on LIBERO while improving zero-shot
generalization on LIBERO-Plus. Real-world experiments further
show improved robustness to unseen visual variations.
These findings support LIT's joint design for reducing
shortcut reliance while retaining the spatial information
needed for action generation.

\bibliographystyle{IEEEtran}
\bibliography{references}

\begin{thebibliography}{10}
\providecommand{\url}[1]{#1}
\csname url@samestyle\endcsname
\providecommand{\newblock}{\relax}
\providecommand{\bibinfo}[2]{#2}
\providecommand{\BIBentrySTDinterwordspacing}{\spaceskip=0pt\relax}
\providecommand{\BIBentryALTinterwordstretchfactor}{4}
\providecommand{\BIBentryALTinterwordspacing}{\spaceskip=\fontdimen2\font plus
\BIBentryALTinterwordstretchfactor\fontdimen3\font minus \fontdimen4\font\relax}
\providecommand{\BIBforeignlanguage}[2]{{%
\expandafter\ifx\csname l@#1\endcsname\relax
\typeout{** WARNING: IEEEtran.bst: No hyphenation pattern has been}%
\typeout{** loaded for the language `#1'. Using the pattern for}%
\typeout{** the default language instead.}%
\else
\language=\csname l@#1\endcsname
\fi
#2}}
\providecommand{\BIBdecl}{\relax}
\BIBdecl

\bibitem{pi05}
K.~Black, N.~Brown, J.~Darpinian, K.~Dhabalia, D.~Driess \emph{et~al.}, ``$\pi_{0.5}$: A vision-language-action model with open-world generalization,'' in \emph{Proceedings of the 9th Conference on Robot Learning (CoRL)}, ser. Proceedings of Machine Learning Research, vol. 305.\hskip 1em plus 0.5em minus 0.4em\relax PMLR, 2025, pp. 17--40.

\bibitem{molmoact2}
H.~Fang, J.~Duan, D.~Clay, S.~Wang, S.~Liu \emph{et~al.}, ``{MolmoAct2}: Action reasoning models for real-world deployment,'' \emph{arXiv preprint arXiv:2605.02881}, 2026.

\bibitem{fastwam}
T.~Yuan, Z.~Dong, Y.~Liu, and H.~Zhao, ``{Fast-WAM}: Do world action models need test-time future imagination?'' \emph{arXiv preprint arXiv:2603.16666}, 2026.

\bibitem{imagewam}
Y.~Zhang, W.~Zhang, Z.~Qi, H.~Zhang, H.~Lin \emph{et~al.}, ``{ImageWAM}: Do world action models really need video generation, or just image editing?'' \emph{arXiv preprint arXiv:2606.19531}, 2026.

\bibitem{diffusionpolicy}
C.~Chi, S.~Feng, Y.~Du, Z.~Xu, E.~Cousineau \emph{et~al.}, ``Diffusion policy: Visuomotor policy learning via action diffusion,'' in \emph{Proceedings of Robotics: Science and Systems (RSS)}, Daegu, Republic of Korea, Jul. 2023.

\bibitem{pi0}
K.~Black, N.~Brown, D.~Driess, A.~Esmail, M.~Equi \emph{et~al.}, ``$\pi_0$: A vision-language-action flow model for general robot control,'' in \emph{Proceedings of Robotics: Science and Systems (RSS)}, Los Angeles, CA, USA, Jun. 2025.

\bibitem{liberoplus}
S.~Fei, S.~Wang, J.~Shi, Z.~Dai, J.~Cai \emph{et~al.}, ``{LIBERO-Plus}: A progressive robustness benchmark for visual-language-action models,'' in \emph{Proceedings of the IEEE/CVF Conference on Computer Vision and Pattern Recognition (CVPR)}, Jun. 2026, pp. 38\,574--38\,583.

\bibitem{shortcutlearning}
R.~Geirhos, J.-H. Jacobsen, C.~Michaelis, R.~Zemel, W.~Brendel \emph{et~al.}, ``Shortcut learning in deep neural networks,'' \emph{Nature Machine Intelligence}, vol.~2, no.~11, pp. 665--673, Nov. 2020.

\bibitem{spatialvla}
D.~Qu, H.~Song, Q.~Chen, Y.~Yao, X.~Ye \emph{et~al.}, ``{SpatialVLA}: Exploring spatial representations for visual-language-action models,'' in \emph{Proceedings of Robotics: Science and Systems (RSS)}, Los Angeles, CA, USA, Jun. 2025.

\bibitem{molmoact}
J.~Lee, J.~Duan, H.~Fang, Y.~Deng, S.~Liu \emph{et~al.}, ``{MolmoAct}: Action reasoning models that can reason in space,'' \emph{arXiv preprint arXiv:2508.07917}, 2025.

\bibitem{tracevla}
R.~Zheng, Y.~Liang, S.~Huang, J.~Gao, H.~Daum{\'e}~III \emph{et~al.}, ``{TraceVLA}: Visual trace prompting enhances spatial-temporal awareness for generalist robotic policies,'' in \emph{International Conference on Learning Representations (ICLR)}, 2025.

\bibitem{la4vla}
T.~Lin, Y.~Du, Y.~Mao, Z.~Ye, Y.~Zhong \emph{et~al.}, ``{LA4VLA}: Learning to act without seeing via language-action pretraining,'' \emph{arXiv preprint arXiv:2606.27295}, 2026.

\bibitem{qwenvla}
Q.~Wang, M.~Li, J.~Guan, J.~Ye, S.~Xie \emph{et~al.}, ``{Qwen-VLA}: Unifying vision-language-action modeling across tasks, environments, and robot embodiments,'' \emph{arXiv preprint arXiv:2605.30280}, 2026.

\bibitem{libero}
B.~Liu, Y.~Zhu, C.~Gao, Y.~Feng, Q.~Liu \emph{et~al.}, ``{LIBERO}: Benchmarking knowledge transfer for lifelong robot learning,'' in \emph{Advances in Neural Information Processing Systems (NeurIPS) Datasets and Benchmarks Track}, vol.~36, 2023.

\bibitem{rtx}
A.~O'Neill \emph{et~al.}, ``Open {X-Embodiment}: Robotic learning datasets and {RT-X} models,'' in \emph{2024 IEEE International Conference on Robotics and Automation (ICRA)}.\hskip 1em plus 0.5em minus 0.4em\relax IEEE, 2024, pp. 6892--6903.

\bibitem{octo}
D.~Ghosh, H.~R. Walke, K.~Pertsch, K.~Black, O.~Mees \emph{et~al.}, ``{Octo}: An open-source generalist robot policy,'' in \emph{Proceedings of Robotics: Science and Systems (RSS)}, Delft, Netherlands, Jul. 2024.

\bibitem{rt2}
B.~Zitkovich, T.~Yu, S.~Xu, P.~Xu, T.~Xiao \emph{et~al.}, ``{RT-2}: Vision-language-action models transfer web knowledge to robotic control,'' in \emph{Proceedings of the 7th Conference on Robot Learning (CoRL)}, ser. Proceedings of Machine Learning Research, vol. 229.\hskip 1em plus 0.5em minus 0.4em\relax PMLR, 2023, pp. 2165--2183.

\bibitem{openvla}
M.~J. Kim, K.~Pertsch, S.~Karamcheti, T.~Xiao, A.~Balakrishna \emph{et~al.}, ``{OpenVLA}: An open-source vision-language-action model,'' in \emph{Proceedings of the 8th Conference on Robot Learning (CoRL)}, ser. Proceedings of Machine Learning Research, vol. 270.\hskip 1em plus 0.5em minus 0.4em\relax PMLR, 2025, pp. 2679--2713.

\bibitem{cogact}
Q.~Li, Y.~Liang, Z.~Wang, L.~Luo, X.~Chen \emph{et~al.}, ``{CogACT}: A foundational vision-language-action model for synergizing cognition and action in robotic manipulation,'' \emph{arXiv preprint arXiv:2411.19650}, 2024.

\bibitem{dreamzero}
S.~Ye, Y.~Ge, K.~Zheng, S.~Gao, S.~Yu \emph{et~al.}, ``World action models are zero-shot policies,'' \emph{arXiv preprint arXiv:2602.15922}, 2026.

\bibitem{knowledgeinsulation}
D.~Driess, J.~T. Springenberg, B.~Ichter, L.~Yu, A.~Li-Bell \emph{et~al.}, ``Knowledge insulating vision-language-action models: Train fast, run fast, generalize better,'' \emph{arXiv preprint arXiv:2505.23705}, 2025.

\bibitem{causalconfusion}
P.~de~Haan, D.~Jayaraman, and S.~Levine, ``Causal confusion in imitation learning,'' in \emph{Advances in Neural Information Processing Systems (NeurIPS)}, vol.~32.\hskip 1em plus 0.5em minus 0.4em\relax Curran Associates, Inc., 2019, pp. 11\,693--11\,704.

\bibitem{simpler}
X.~Li, K.~Hsu, J.~Gu, O.~Mees, K.~Pertsch \emph{et~al.}, ``Evaluating real-world robot manipulation policies in simulation,'' in \emph{Proceedings of the 8th Conference on Robot Learning (CoRL)}, ser. Proceedings of Machine Learning Research, vol. 270.\hskip 1em plus 0.5em minus 0.4em\relax PMLR, 2025, pp. 3705--3728.

\bibitem{vlabench}
S.~Zhang, Z.~Xu, P.~Liu, X.~Yu, Y.~Li \emph{et~al.}, ``{VLABench}: A large-scale benchmark for language-conditioned robotics manipulation with long-horizon reasoning tasks,'' in \emph{Proceedings of the IEEE/CVF International Conference on Computer Vision (ICCV)}, Oct. 2025, pp. 11\,142--11\,152.

\bibitem{r3m}
S.~Nair, A.~Rajeswaran, V.~Kumar, C.~Finn, and A.~Gupta, ``{R3M}: A universal visual representation for robot manipulation,'' in \emph{Proceedings of the 6th Conference on Robot Learning (CoRL)}, ser. Proceedings of Machine Learning Research, vol. 205.\hskip 1em plus 0.5em minus 0.4em\relax PMLR, 2023, pp. 892--909.

\bibitem{mvp}
T.~Xiao, I.~Radosavovic, T.~Darrell, and J.~Malik, ``Masked visual pre-training for motor control,'' \emph{arXiv preprint arXiv:2203.06173}, 2022.

\bibitem{ecot}
M.~Zawalski, W.~Chen, K.~Pertsch, O.~Mees, C.~Finn, and S.~Levine, ``Robotic control via embodied chain-of-thought reasoning,'' in \emph{Proceedings of the 8th Conference on Robot Learning (CoRL)}, ser. Proceedings of Machine Learning Research, vol. 270.\hskip 1em plus 0.5em minus 0.4em\relax PMLR, 2025, pp. 3157--3181.

\bibitem{cotvla}
Q.~Zhao, Y.~Lu, M.~J. Kim, Z.~Fu, Z.~Zhang \emph{et~al.}, ``{CoT-VLA}: Visual chain-of-thought reasoning for vision-language-action models,'' in \emph{Proceedings of the IEEE/CVF Conference on Computer Vision and Pattern Recognition (CVPR)}, Jun. 2025, pp. 1702--1713.

\bibitem{hamster}
Y.~Li, Y.~Deng, J.~Zhang, J.~Jang, M.~Memmel \emph{et~al.}, ``{HAMSTER}: Hierarchical action models for open-world robot manipulation,'' in \emph{International Conference on Learning Representations (ICLR)}, 2025.

\bibitem{hamster3d}
D.~Hwang, B.~Lee, D.~Kim, H.~Jang, H.~Jin \emph{et~al.}, ``{3D HAMSTER}: Bridging planning and control in hierarchical vision language action models through {3D} trajectory guidance,'' in \emph{IEEE/RSJ International Conference on Intelligent Robots and Systems (IROS)}, 2026.

\bibitem{crosswaydiffusion}
X.~Li, V.~Belagali, J.~Shang, and M.~S. Ryoo, ``Crossway diffusion: Improving diffusion-based visuomotor policy via self-supervised learning,'' in \emph{2024 IEEE International Conference on Robotics and Automation (ICRA)}.\hskip 1em plus 0.5em minus 0.4em\relax IEEE, 2024, pp. 16\,841--16\,849.

\bibitem{oreo}
J.~Park, Y.~Seo, C.~Liu, L.~Zhao, T.~Qin \emph{et~al.}, ``Object-aware regularization for addressing causal confusion in imitation learning,'' in \emph{Advances in Neural Information Processing Systems (NeurIPS)}, vol.~34.\hskip 1em plus 0.5em minus 0.4em\relax Curran Associates, Inc., 2021, pp. 3029--3042.

\bibitem{selectivevisual}
A.~Eftekhar, K.-H. Zeng, J.~Duan, A.~Farhadi, A.~Kembhavi, and R.~Krishna, ``Selective visual representations improve convergence and generalization for embodied {AI},'' in \emph{International Conference on Learning Representations (ICLR)}, 2024.

\bibitem{jing2026learning}
D.~Jing, T.~Zhang, J.~Liu, J.~Zhao, Z.~Sun, L.~E. Li, Z.~Lu, and M.~Ding, ``Learning action priors for cross-embodiment robot manipulation,'' \emph{arXiv preprint arXiv:2606.26095}, 2026.

\bibitem{apt}
K.~Xu, Z.~Zhu, A.~Chen, R.~Xiong, and Y.~Wang, ``{APT}: Action expert pretraining improves instruction generalization of vision-language-action policies,'' \emph{arXiv preprint arXiv:2606.12366}, 2026.

\bibitem{flowmatching}
Y.~Lipman, R.~T.~Q. Chen, H.~Ben-Hamu, M.~Nickel, and M.~Le, ``Flow matching for generative modeling,'' in \emph{International Conference on Learning Representations (ICLR)}, 2023.

\bibitem{wang2026vlaadapter}
Y.~Wang, P.~Ding, L.~Li, C.~Cui, Z.~Ge \emph{et~al.}, ``{VLA-Adapter}: An effective paradigm for tiny-scale vision-language-action model,'' in \emph{Proceedings of the AAAI Conference on Artificial Intelligence}, vol.~40, no.~22, 2026.

\end{thebibliography}

\end{document}